\pdfoutput=1

\documentclass[11pt]{article}

\usepackage[]{emnlp2021}

\usepackage{times}
\usepackage{latexsym}

\usepackage[T1]{fontenc}

\usepackage[utf8]{inputenc}

\usepackage{microtype}

\usepackage{amsfonts}
\usepackage{amsmath}
\usepackage{graphicx}
\usepackage{multirow}
\usepackage{caption}
\usepackage{subfigure}
\usepackage{enumitem}
\usepackage{float}
\usepackage{makecell}
\usepackage{booktabs}
\usepackage{tabu}
\usepackage{tabularx}
\usepackage{pifont}

\title{Data Augmentation for Cross-Domain Named Entity Recognition}

\author{
Shuguang Chen \textsuperscript{$\dagger$},
Gustavo Aguilar \textsuperscript{$\dagger$},
Leonardo Neves \textsuperscript{$\ast$} \and 
Thamar Solorio \textsuperscript{$\dagger$} \\
University of Houston \textsuperscript{$\dagger$} \\
Snap Inc. \textsuperscript{$\ast$}\\
\{schen52, gaguilaralas, tsolorio\}@uh.edu\textsuperscript{$\dagger$} \\ 
lneves@snap.com\textsuperscript{$\ast$}
}

\begin{document}
\maketitle

\begin{abstract}
Current work in named entity recognition (NER) shows that data augmentation techniques can produce more robust models. However, most existing techniques focus on augmenting in-domain data in low-resource scenarios where annotated data is quite limited. In contrast, we study cross-domain data augmentation for the NER task. We investigate the possibility of leveraging data from high-resource domains by projecting it into the low-resource domains. Specifically, we propose a novel neural architecture to transform the data representation from a high-resource to a low-resource domain by learning the patterns (e.g. style, noise, abbreviations, etc.) in the text that differentiate them and a shared feature space where both domains are aligned. We experiment with diverse datasets and show that transforming the data to the low-resource domain representation achieves significant improvements over only using data from high-resource domains. \footnote{We release the code at \url{https://github.com/RiTUAL-UH/style_NER}.}
\end{abstract}

\section{Introduction}
Named entity recognition (NER) has seen significant performance improvements with the recent advances of pre-trained language models \citep{akbik-etal-2019-pooled, devlin-etal-2019-bert}. However, the high performance of such models usually relies on the size and quality of training data. When used under low-resource or even zero-resource scenarios, those models struggle to generalize over diverse domains \citep{Fu_Liu_Zhang_2020}, and the performance drops dramatically due to the lack of annotated data. Unfortunately, annotating more data is often expensive and time-consuming, and it requires expert domain knowledge. Moreover, annotated data can quickly become outdated in domains where language changes rapidly (e.g, social media), leading to the temporal drift problem \citep{rijhwani-preotiuc-pietro-2020-temporally}.

A common approach to alleviate the limitations mentioned above is data augmentation, where automatically generated data can increase the size and diversity in the training set, while resulting in model performance gains. But data augmentation in the context of NER is still understudied. Approaches that directly modify words in the training set (e.g, synonym replacement \citep{Zhang2015CharacterlevelCN} and word swap \citep{wei-zou-2019-eda}) can inadvertently result in incorrectly labeled entities after modification. Recent work in NER for low resource scenarios is promising \citep{dai-adel-2020-analysis, ding-etal-2020-daga} but it is limited to same domain settings and improvements decrease drastically with smaller sizes of training data.

\begin{figure}[t]
\centering
\subfigure{
    \begin{minipage}[t]{\linewidth}
    \centering
    \includegraphics[width=1\linewidth]{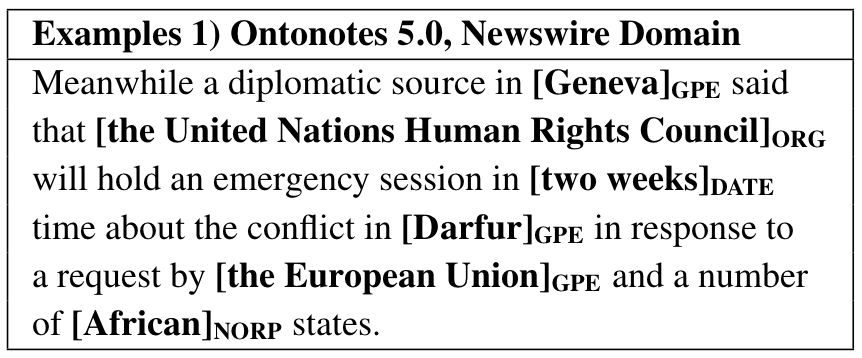}
    \end{minipage}
}
\subfigure{
    \begin{minipage}[t]{\linewidth}
    \centering
    \includegraphics[width=1\linewidth]{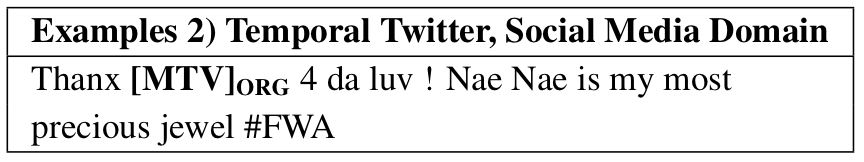}
    \end{minipage}
}
\caption{Examples from Ontonotes 5.0 dataset and Temporal Twitter dataset. The language variations and abbreviations in the text from social media domain make it clearly different from the formal text in newswire domain. }
\label{fig: examples}
\vspace{-0.5cm}
\end{figure}
To facilitate research in this direction, we investigate leveraging data from high-resource domains by projecting it into low-resource domains. Based on our observations, the text in different domains usually presents unique patterns (e.g. style, noise, abbreviations, etc.). As shown in Figure \ref{fig: examples}, the text in the newswire domain is long and formal, while the text in the social media domain is short and noisy, often presenting many grammar errors, spelling mistakes, and language variations. In this work, we hypothesize that even though the textual patterns are different across domains, the semantics of text are still transferable. Additionally, there are some invariables in the way the named entities appear and we assume that the model can learn from them. In this work, we introduce a cross-domain autoencoder model capable of extracting the textual patterns in different domains and learning a shared feature space where domains are aligned. We evaluate our data augmentation method by conducting experiments on two datasets, including six different domains and ten domain pairs, showing that transforming the data from high-resource to low-resource domains is a more powerful method than simply using the data from high-resource domains. We also explore our data augmentation approach in the context of the NER task in low-resource scenarios for both in-domain and out-of-domain data.

To summarize, we make the following contributions:
\begin{enumerate}[topsep=0pt,itemsep=-1ex,partopsep=1ex,parsep=1ex]
\item We propose a novel neural architecture that can learn the textual patterns and effectively transform the text from a high-resource to a low-resource domain.
\item We systematically evaluate our proposed method on two datasets, including six different domains and ten different domain pairs, and show the effectiveness of cross-domain data augmentation for the NER task.
\item We empirically explore our approach in low-resource scenarios and expose the case where our approach could benefit the low-resource NER task
\end{enumerate}{}

\section{Related work}
Data augmentation aims to increase the size of training data by slightly modifying the copies of already existing data or adding newly generated synthetic data from existing data \citep{hou-etal-2018-sequence, wei-zou-2019-eda}. It has become more practical for NLP tasks in recent years, especially in low-resource scenarios where annotated data is limited \citep{fadaee-etal-2017-data, xia-etal-2019-generalized}. Without collecting new data, this technique reduces the cost of annotation and boosts the model performance.

Previous work has studied the data augmentation for both token-level tasks \citep{sahin-steedman-2018-data, gao-etal-2019-soft} and sequence-level tasks \citep{wang-yang-2015-thats, min-etal-2020-syntactic}. Related to data augmentation on NER, \citet{dai-adel-2020-analysis} conducted a study that primarily focuses on the simple data augmentation methods such as synonym replacement (i.e., replace the token with one of its synonyms) and mention replacement (i.e., randomly replace the mention with another one that has the same entity type with the replacement). \citet{zhang-etal-2020-seqmix} studied sequence mixup (i.e., mix eligible sequences in the feature space and the label space) to improve the data diversity and enhance sequence labeling for active learning. \citet{ding-etal-2020-daga} presented a novel approach using adversarial learning to generate high-quality synthetic data, which is applicable to both supervised and semi-supervised settings. 

In cross-domain settings, NER models struggle to generalize over diverse genres \citep{rijhwani-preotiuc-pietro-2020-temporally, Fu_Liu_Zhang_2020}. Most existing work mainly studies domain adaptation \citep{liu-etal-2020-zero, jia-etal-2019-cross, wang-etal-2020-multi-domain-named, liu2020crossner} which aims to adapt a neural model from a source domain to achieve better performance on the data from the target domain. \citet{liu-etal-2020-zero} proposed a zero-resource cross-domain framework to learn the general representations of named entities.
\citet{jia-etal-2019-cross} studied the knowledge of domain difference and presented a novel parameter generation network. Other efforts include the different domain adaptation settings \citep{wang-etal-2020-multi-domain-named} and effective cross-domain evaluation \citep{liu2020crossner}. In our work, we focus on cross-domain data augmentation. The proposed method aims to map data from a high-resource domain to a low-resource domain. By learning the textual patterns of the data from different domains, our proposed method transform the data from one domain to another and boosts the model performance with the generated data for NER in low-resource settings.

\section{Proposed Method}
\begin{figure*}[ht]
\centering
\subfigure[Denoising Reconstruction]{
    \begin{minipage}[t]{\linewidth}
    \centering
    \includegraphics[width=0.75\linewidth]{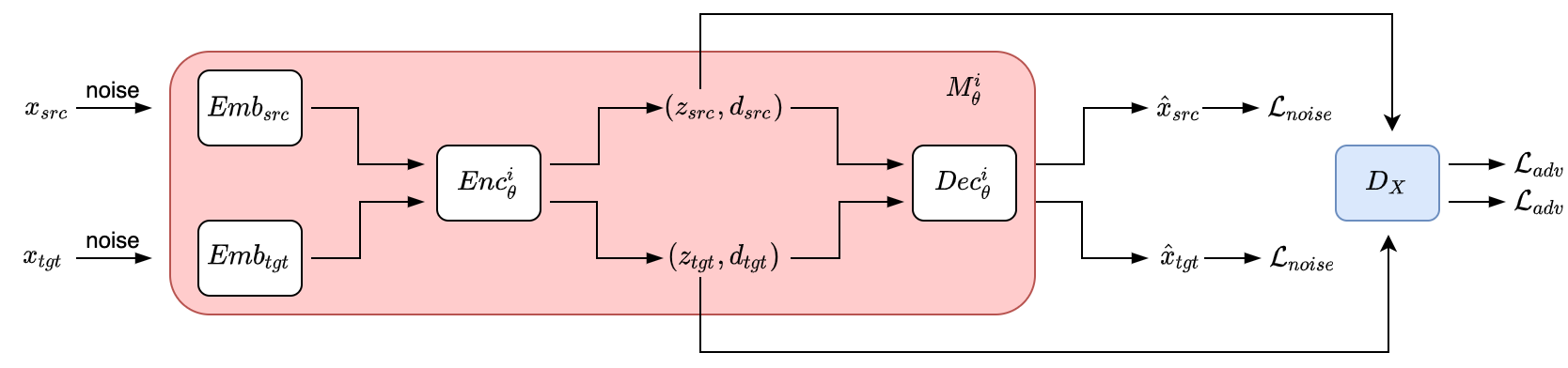}
    \end{minipage}
    \label{fig: denoising_reconstruction}
}
\subfigure[Detransforming Reconstruction]{
    \begin{minipage}[t]{\linewidth}
    \centering
    \includegraphics[width=1\linewidth]{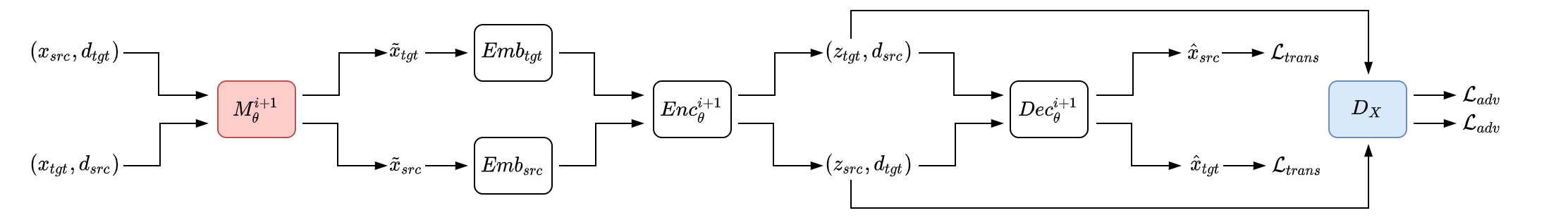}
    \end{minipage}
    \label{fig: detransforming_reconstruction}
}
\caption{The general architecture of our proposed method. (a) Figure shows the architecture for denoising reconstruction, which aims to reconstruct each input sentence from its noisy version in its corresponding domain. (b) Figure shows the details of reconstructing each input sentence from its transformed version in its corresponding domain. We call this detransforming reconstruction.}
\label{fig: overall_architecture}
\end{figure*}
In this work, we propose a novel neural architecture to augment the data by transforming the text from a high-resource domain to a lower-resource domain for the NER task. The overall neural architecture is shown in Figure \ref{fig: overall_architecture}. 

We consider two unparalleled datasets: one from the source domain $D_{src}$ and one from the target domain $D_{tgt}$. We first linearize all sentences by inserting every entity label before the corresponding word. At each iteration, we randomly pair a sentence from $D_{src}$ and a sentence from $D_{tgt}$ as the input to the model. The model starts with word-by-word denoising reconstruction and then detransforming reconstruction. In denoising reconstruction, we aim to train the model to learn a compressed representation of an input based on the domain it comes from in an unsupervised way. We inject noise into each input sentence by shuffling, dropping, or masking some words. The encoder is trained to capture the textual semantics and learn the pattern that makes each sentence different from sentences in other domains. Then we train the decoder by minimizing a training objective that measures its ability to reconstruct each sentence from its noisy version in its corresponding domain. In detransforming reconstruction, the goal is to transform sentences from one domain to another domain based on their textual semantics. We first transform each sentence from the source/target domain to the target/source domain with the model from the previous training step as the input. The encoder then generates latent representations for transformed sentences. After that, different from denoising reconstruction, the decoder here is trained to reconstruct each sentence from its transformed version in its corresponding domain. Besides denoising and detransforming reconstruction, we also train a discriminator to distinguish whether the latent vector generated by the encoder is from the source domain or target domain. In this case, the encoder can generate a meaningful intermediate representation. Otherwise, the model would bypass the intermediate mapping step between domains and replace it by memorizing rather than generalizing. In the following sections, we will introduce the details of our model architecture and the training algorithm.

\subsection{Data Pre-processing}
\label{ref: data_preprocessing}
Following \citet{ding-etal-2020-daga}, we perform sentence linearization so that the model can learn the distribution and the relationship of words and labels. In this work, we use the standard BIO schema \citep{tjong-kim-sang-veenstra-1999-representing}. Given a sequence of words $w = \left \{w_{1}, w_{2}, ..., w_{n}\right \}$ and a sequence of labels $l = \left \{l_{1}, l_{2}, ..., l_{n}\right \}$, we first linearize the words with labels by putting every label $l_{i}$ before the corresponding word $w_{i}$. Then we generate a new sentence $x = \left \{l_{1}, w_{1}, l_{2}, w_{2}, ..., l_{n}, w_{n}\right \}$ and drop all $O$ labels from them as the input. Special tokens \texttt{<BOS>} and \texttt{<EOS>} are inserted at the beginning and the end of each input sentence. 

\subsection{Cross-domain Autoencoder}
\label{ref: cross_domain_language_model}

\paragraph{Word-level Robustness}
\label{ref: word_level_robustness}
Our cross-domain autoencoder model involves an encoder $Enc: x \rightarrow z$ that maps input sequences from data space to latent space. Previous work \citep{Shen2020EducatingTA} has demonstrated that input perturbations are particularly useful for discrete text modeling with powerful sequence networks, as they encourage the preservation of data structure in latent space representations. In this work, we perturb each input sentence by injecting noise with three different operations (see Table \ref{ref: word_operations}) to ensure that similar input sentences can have similar latent representations.
\begin{table}[t]
\small
    \centering
    \renewcommand{\arraystretch}{1.2}
        \resizebox{\linewidth}{!}{
            \begin{tabular}{ll}
                \toprule
                \bf Operation & \bf Description \\
                \toprule
                Shuffle & generate a new permutation of all words\\
                Dropout & randomly drop a word from the sequence\\
                Mask    & randomly mask a word with \texttt{<MSK>} token\\
                \toprule
            \end{tabular}
        }
    \caption{Word-level operations to inject noise in each input sequence. Each operation is randomly applied to input sequences with a certain probability $p$.}
    \label{ref: word_operations}
\end{table}  

\paragraph{Denoising Reconstruction}
\label{ref: denoising_reconstruction}
The neural architecture for denoising reconstruction is shown in Figure \ref{fig: denoising_reconstruction}. Consider a pair of two unparalleled sentences: one sentence $x_{src}$ from $D_{src}$ in the source domain and another sentence $x_{tgt}$ from $D_{tgt}$ in the target domain. The model is trained to reconstruct each sentence by sharing the same encoder and decoder parameters while using different embedding lookup tables. The token embedders $Emb_{src}$ and $Emb_{tgt}$ hold a lookup table of the corresponding domains. The encoder is a bi-directional LSTM model that takes the noisy linearized sentences as input and returns hidden states as latent vectors. At each decoding step, the decoder takes the current word and the latent vector from the previous step as input. Then it uses the vocabulary in the corresponding domain to project each vector from latent space to vocabulary space and predicts the next word with additive attention \cite{Bahdanau2015NeuralMT}.

The training objective for denoising reconstruction is defined as below. The goal of this training objective is to force the model to learn a shared space where both domains are aligned through the latent vectors and generate a compressed version of input sentence.
\begin{equation*}
\left.\begin{aligned}
& \mathcal{L}_{noise}(\hat{x}, x) = -\sum_{i=1}^{N}  x_{i} \cdot  log{\hat{x}_{i}}\\
\end{aligned}\right.
\end{equation*}

\paragraph{Detransforming Reconstruction}
\label{ref: detransforming_reconstruction}
In detransforming reconstruction, the first step is to transform each sentence from the source/target domain to the target/source domain. As shown in Figure \ref{fig: detransforming_reconstruction}, given a pair of sequences $x_{src}$ and $x_{tgt}$ from the source and target domain, we first map $x_{src}$ to $\tilde{x}_{tgt}$ in the target domain and $x_{tgt}$ to $\tilde{x}_{src}$ in the source domain by applying the model $M^{i-1}_{\theta}$, which includes embedders, encoder, and decoder, from previous training step. After that, we feed $\tilde{x}_{tgt}$ and $\tilde{x}_{src}$ to the encoder and generate compressed latent representations $z_{tgt}$ and $z_{src}$. Then the decoder maps $z_{tgt}$ to $\hat{x}_{src}$ in the source domain and $z_{src}$ to $\hat{x}_{tgt}$ in the target domain. The goal is to learn the mapping between different domains and reconstruct a sequence from its transformed version in its corresponding domain.

The training objective for detransforming reconstruction is shown below. 
\begin{equation*}
\left.\begin{aligned}
& \mathcal{L}_{trans}(\hat{x}, x) = -\sum_{i=1}^{N}  x_{i} \cdot log{\hat{x}_{i}}\\
\end{aligned}\right.
\end{equation*}

\paragraph{Domain Classification}
\label{ref: domain_classification}
For domain classification, we apply adversarial training. We use the encoder to extract the textual patterns of sentences from different domains. The encoder generates the latent representations for the noised or transformed version of inputs and the discriminator tells if the given latent vector is actually from the source or target domain. Then the encoder will refine its technique to fool the discriminator in a way that will end up capturing the patterns to convert text from the source/target domain to the target/source domain. The discriminator is first trained in the denoising reconstruction and then fine-tuned in the detransforming reconstruction to distinguish source domain sentences and target domain sentences. As shown in Figure \ref{fig: overall_architecture}, the discriminator $D_{X}$ takes inputs from both domains without knowing where the sequences come from. Then, the model predicts the corresponding domains of the inputs. The inputs are the latent vectors $z$, where both domains have been mapped to the same space. We formulate this task as a binary classification task. The training objective of adversarial training is described as below:
\begin{equation*}
\left.\begin{aligned}
\mathcal{L}_{adv}(\hat{z}_{i}, z_{i}) & = \!-\!\sum_{i=1}^{N} z_{i}log{\hat{z}_{i}} + (1 - z_{i})log{(1 - \hat{z}_{i})} \\
\end{aligned}\right.
\end{equation*}

\paragraph{Final Training Objective}
The final training objective is defined as the weighted sum of $\mathcal{L}_{noise}$, $\mathcal{L}_{trans}$, and $\mathcal{L}_{adv}$:
\begin{equation*}
\left.\begin{aligned}
\mathcal{L}_{final}(\theta) & = \lambda_{1}\mathcal{L}_{noise} + \lambda_{2}\mathcal{L}_{trans} + \lambda_{3}\mathcal{L}_{adv}\\
\end{aligned}\right.
\end{equation*}

where $\lambda_{1}$, $\lambda_{2}$, and $\lambda_{3}$ are parameters that weight the importance of each loss.

\subsection{Training Algorithm}
\label{ref: training_algorithm}
Based on our observation, the model's ability to reconstruct sentences across domains highly relies on the denoising reconstruction and domain classification components. Therefore, in this work, we take two phases to train our model. In the first phase, we train the model with only denoising reconstruction and domain classification so that it can learn the textual pattern and generate compressed representations of the data from each domain. We calculate the perplexity for denoising reconstruction as the criterion to select the best model across iterations. In the second phase, we train the model together with denoising reconstruction, detransforming reconstruction, and the domain classification. The goal is to align the compressed representations of the data from different domains so that the model can project the data from one domain to another. We calculate the sum of the perplexity for both denoising and detransforming reconstruction as the model selection criterion. 

\subsection{Data Post-processing}
\label{ref: data_postprocessing}
We generate synthetic data using the cross-domain autoencoder model as described in Section \ref{ref: cross_domain_language_model}. We convert the generated data from the linearized format to the same format as gold data. We use the following rules to post-process the generated data: 1) remove sequences that do not follow the standard BIO schema; 2) remove sequences that have <UNK> or <MSK> tokens; 3) remove sequences that do not have any entity tags.

\section{Experiments}
In this section, we will introduce the cross-domain mapping experiment and the NER experiment. In the cross-domain mapping experiment, we analyze the reconstruction and generation capability of the proposed model. We then tested our proposed method and evaluated the data generated from our model on the NER task. Details of the data set, experimental setup, and results are described below.

\subsection{Datasets}
In our experiments, we use two datasets: Ontonotes 5.0 Dataset \citep{pradhan-etal-2013-towards} and Temporal Twitter Dataset \citep{rijhwani-preotiuc-pietro-2020-temporally}. We select data from six different domains in English language, including \textit{Broadcast  Conversation (BC)}, \textit{Broadcast News (BN)}, \textit{Magazine (MZ)}, \textit{Newswire (NW)}, \textit{Web Data (WB)}, and \textit{Social Media (SM)}. All the data is annotated with the following 18 entity tags: \textit{PERSON}, \textit{NORP}, \textit{FAC}, \textit{ORG}, \textit{GPE}, \textit{LOC}, \textit{PRODUCT}, \textit{EVENT}, \textit{WORK\_OF\_ART}, \textit{LAW}, \textit{LANGUAGE}, \textit{DATE}, \textit{TIME}, \textit{PERCENT}, \textit{MONEY}, \textit{QUANTITY}, \textit{ORDINAL}, \textit{CARDINAL}. Below we describe how we pre-process each dataset:

\paragraph{Ontonotes 5.0 Dataset} 
We use subsets from five different domains, including \textit{Broadcast  Conversation (BC)}, \textit{Broadcast News (BN)}, \textit{Magazine (MZ)}, \textit{Newswire (NW)}, and \textit{Web Data (WB)}. Following \citet{pradhan-etal-2013-towards}, we use the same splits and remove the repeated sequences from each dataset.

\paragraph{Temporal Twitter Dataset} 
This dataset was collected from \textit{Social Media (SM)} domain. It includes tweets from 2014 to 2019, with 2,000 samples from each year. We use the data from 2014 to 2018 as the training set. Following \citet{rijhwani-preotiuc-pietro-2020-temporally}, we use 500 samples from 2019 as the validation set and another 1,500 samples from 2019 as the test set.

\subsection{Cross-domain Mapping}
In this section, we describe the experimental settings of our proposed cross-domain autoencoder model and report the evaluation results. 

\paragraph{Cross-domain Autoencoder}
We use our proposed cross-domain autoencoder model (described in Section \ref{ref: cross_domain_language_model}) to generate synthetic data. In our experiments, we build the vocabulary with the most common 10K words and 5 special tokens: \texttt{<PAD>}, \texttt{<UNK>}, \texttt{<BOS>}, \texttt{<EOS>} and \texttt{<MSK>}. We use a bi-directional LSTM layer as the encoder and a LSTM layer as the decoder. For the discriminator, we use a linear layer. The hyper-parameters are described in Appendix \ref{sec: experimental_settings}.

\paragraph{Results} 
For cross-domain mapping experiments, we consider two different domains as the source domain: \textit{NW} and \textit{SM}. The textual patterns in \textit{NW} are similar to that in other domains while the textual patterns in \textit{SM} is quite different from that in other domains (see Appendix \ref{sec: domain_similarity} on domain similarity). In Table \ref{ref: cross_domain_mapping_results}, we report the results of cross-domain mapping experiments on ten different domain pairs. We use perplexity as the metric to evaluate reconstruction. The lower perplexity indicates a higher accuracy and a higher quality of reconstruction. From the results of our experiments, we notice that the average perplexity with \textit{NW} as source domain is lower than the average perplexity with \textit{SM} as source domain, indicating that the model can easily reconstruct both in-domain and out-of-domain sentences when the textual patterns are transferable. 
\begin{table}[t]
\small
    \centering
    \renewcommand{\arraystretch}{1.2}
    \resizebox{0.75\linewidth}{!}{
        \begin{tabular}{l|cc|rr}
            \toprule
            \multirow{2}{*}{\bf Exp ID} & \multirow{2}{*}{\bf Source} & \multirow{2}{*}{\bf Target} & \multicolumn{2}{c}{\bf Reconstruction}\\ 
            \cmidrule(lr){4-5}
            & &  & \bf Dev & \bf Test \\
            \toprule
            Exp 1.0 & NW & BC   & 15.61 & 14.94\\
            Exp 1.1 & NW & BN   & 5.43  & 4.31 \\
            Exp 1.2 & NW & MZ   & 6.72  & 5.98 \\
            Exp 1.3 & NW & WB   & 3.84  & 3.73 \\
            Exp 1.4 & NW & SM   & 8.31  & 7.65 \\
            \toprule
            Exp 1.5 & SM & BC   & 16.02 & 14.67\\
            Exp 1.6 & SM & BN   & 11.34 & 12.58\\
            Exp 1.7 & SM & MZ   & 14.64 & 15.28\\
            Exp 1.8 & SM & NW   & 8.31  & 7.65 \\
            Exp 1.9 & SM & WB   & 9.08  & 8.54 \\
            \toprule
        \end{tabular}
    }
    \caption{The results of our cross-domain autoencoder model on each domain pair. The scores for reconstruction, including both denoising and detransforming reconstruction, are calculated with the perplexity metric.}
    \label{ref: cross_domain_mapping_results}
\end{table}

\subsection{Named Entity Recognition}
\begin{table}[t]
\small
    \centering
    \renewcommand{\arraystretch}{1.2}
    \resizebox{0.95\linewidth}{!}{
    \begin{tabular}{l|l|llll|r}
    \toprule
    \multirow{2}{*}{\bf Domain Pair} & \multirow{2}{*}{\bf Data} & \multicolumn{4}{c|}{\bf Number of Training Samples} & \multirow{2}{*}{\bf Gain}\\ 
    \cmidrule(lr){3-6}
    & & \bf 1K & \bf 2K & \bf 3K & \bf 4K & \\
    \toprule
    \toprule
    \multirow{3}{*}{NW $\rightarrow$ BC}& NW        & 66.53 & 69.70 & 72.47 & 73.05& - \\
                                        & Gen       & 59.14 {\color{red} \makebox[0pt][l]{$\downarrow$} } & 62.32 {\color{red} \makebox[0pt][l]{$\downarrow$} } & 64.49 {\color{red} \makebox[0pt][l]{$\downarrow$} } & 65.75 {\color{red} \makebox[0pt][l]{$\downarrow$} }& {\color{red}  -7.51 } \\
                                        & BC        & 69.65 & 75.13 & 78.44 & 79.45& - \\
    \toprule
    \multirow{3}{*}{NW $\rightarrow$ BN}& NW        & 77.55 & 80.77 & 82.30 & 83.36& - \\
                                        & Gen       & 78.31 {\color{blue} \makebox[0pt][l]{$\uparrow$} } & 81.29 {\color{blue} \makebox[0pt][l]{$\uparrow$} } & 82.32 {\color{blue} \makebox[0pt][l]{$\uparrow$} } & 83.48 {\color{blue} \makebox[0pt][l]{$\uparrow$} }& {\color{blue}  +0.36 } \\
                                        & BN        & 82.63 & 86.26 & 87.49 & 88.72& - \\
    \toprule
    \multirow{3}{*}{NW $\rightarrow$ MZ}& NW        & 71.41 & 73.61 & 75.14 & 76.62& - \\
                                        & Gen       & 72.16 {\color{blue} \makebox[0pt][l]{$\uparrow$} } & 74.64 {\color{blue} \makebox[0pt][l]{$\uparrow$} } & 75.99 {\color{blue} \makebox[0pt][l]{$\uparrow$} } & 77.31 {\color{blue} \makebox[0pt][l]{$\uparrow$} }& {\color{blue}  +0.83 } \\
                                        & MZ        & 82.63 & 83.54 & 85.55 & 86.05& - \\
    \toprule
    \multirow{3}{*}{NW $\rightarrow$ WB}& NW        & 41.95 & 43.88 & 44.52 & 45.35& - \\
                                        & Gen       & 43.25 {\color{blue} \makebox[0pt][l]{$\uparrow$} } & 44.42 {\color{blue} \makebox[0pt][l]{$\uparrow$} } & 45.10 {\color{blue} \makebox[0pt][l]{$\uparrow$} } & 45.61 {\color{blue} \makebox[0pt][l]{$\uparrow$} }& {\color{blue}  +0.67 } \\
                                        & WB        & 46.22 & 55.47 & 58.77 & 60.37& - \\
    \toprule
    \multirow{3}{*}{NW $\rightarrow$ SM}& NW        & 34.71 & 35.47 & 35.69 & 35.69& - \\
                                        & Gen       & 43.19 {\color{blue} \makebox[0pt][l]{$\uparrow$} } & 43.85 {\color{blue} \makebox[0pt][l]{$\uparrow$} } & 44.29 {\color{blue} \makebox[0pt][l]{$\uparrow$} } & 44.82 {\color{blue} \makebox[0pt][l]{$\uparrow$} }& {\color{blue}  +8.65 } \\
                                        & SM        & 69.91 & 73.33 & 73.99 & 74.56& - \\
    \toprule
    \toprule
    \multirow{3}{*}{SM $\rightarrow$ BC}& SM        & 26.94 & 28.87 & 29.61 & 29.80& - \\
                                        & Gen       & 36.59 {\color{blue} \makebox[0pt][l]{$\uparrow$} } & 37.08 {\color{blue} \makebox[0pt][l]{$\uparrow$} } & 38.40 {\color{blue} \makebox[0pt][l]{$\uparrow$} } & 38.76 {\color{blue} \makebox[0pt][l]{$\uparrow$} }& {\color{blue}  +8.90 } \\
                                        & BC        & 69.65 & 75.13 & 78.44 & 79.45& - \\
    \toprule
    \multirow{3}{*}{SM $\rightarrow$ BN}& SM        & 28.63 & 29.90 & 30.45 & 30.83& - \\
                                        & Gen       & 47.28 {\color{blue} \makebox[0pt][l]{$\uparrow$} } & 48.32 {\color{blue} \makebox[0pt][l]{$\uparrow$} } & 49.15 {\color{blue} \makebox[0pt][l]{$\uparrow$} } & 50.79 {\color{blue} \makebox[0pt][l]{$\uparrow$} }& {\color{blue}  +18.93 } \\
                                        & BN        & 82.63 & 86.26 & 87.49 & 88.72& - \\
    \toprule
    \multirow{3}{*}{SM $\rightarrow$ MZ}& SM        & 20.54 & 25.76 & 27.48 & 28.89& - \\
                                        & Gen       & 41.11 {\color{blue} \makebox[0pt][l]{$\uparrow$} } & 42.78 {\color{blue} \makebox[0pt][l]{$\uparrow$} } & 44.12 {\color{blue} \makebox[0pt][l]{$\uparrow$} } & 45.89 {\color{blue} \makebox[0pt][l]{$\uparrow$} }& {\color{blue}  +17.81 } \\
                                        & MZ        & 82.11 & 83.54 & 85.55 & 86.05& - \\
    \toprule
    \multirow{3}{*}{SM $\rightarrow$ NW}& SM        & 25.94 & 28.83 & 29.34 & 30.20& - \\
                                        & Gen       & 35.98 {\color{blue} \makebox[0pt][l]{$\uparrow$} } & 38.33 {\color{blue} \makebox[0pt][l]{$\uparrow$} } & 39.55 {\color{blue} \makebox[0pt][l]{$\uparrow$} } & 40.84 {\color{blue} \makebox[0pt][l]{$\uparrow$} }& {\color{blue}  +10.10 } \\
                                        & NW        & 79.50 & 82.81 & 85.55 & 86.49& - \\
    \toprule
    \multirow{3}{*}{SM $\rightarrow$ WB}& SM        & 25.70 & 26.23 & 26.41 & 26.52& - \\
                                        & Gen       & 36.15 {\color{blue} \makebox[0pt][l]{$\uparrow$} } & 36.73 {\color{blue} \makebox[0pt][l]{$\uparrow$} } & 37.65 {\color{blue} \makebox[0pt][l]{$\uparrow$} } & 38.13 {\color{blue} \makebox[0pt][l]{$\uparrow$} }& {\color{blue}  +10.95 } \\
                                        & WB        & 46.22 & 55.47 & 58.77 & 60.37& - \\
    \toprule
    \toprule
    \end{tabular}
    }
    \caption{The results of our proposed data augmentation for the NER task on ten different domain pairs. Scores are calculated with the micro F1 metric. Note that there is a decreasing tendency ({\color{red} $\downarrow$}) for the pairs with similar textual pattern; The scores from the pairs with dissimilar textual patterns tend to increase ({\color{blue} $\uparrow$}).}
    \label{ref: named_entity_recognition_results}
\end{table}
\begin{table*}[ht]
\small
    \centering
    \renewcommand{\arraystretch}{1.2}
    \resizebox{0.9\linewidth}{!}{
        \begin{tabular}{l|l|llll|llll}
        \toprule
        \multirow{3}{*}{\bf Exp ID} & \multirow{3}{*}{\bf Method} & \multicolumn{4}{c|}{\bf NW $\rightarrow$ SM} & \multicolumn{4}{c}{\bf SM $\rightarrow$ NW}\\
        \cmidrule(lr){3-10}
            & & \multicolumn{4}{c|}{\bf Number of Training Samples} & \multicolumn{4}{c}{\bf Number of Training Samples}\\ 
        \cmidrule(lr){3-6} \cmidrule(lr){7-10}
        & & \bf 1K & \bf 2K & \bf 3K & \bf 4K & \bf 1K & \bf 2K & \bf 3K & \bf 4K\\
        \toprule
        Exp 2.0 & Baseline (No Augmentation)        & 34.71 & 35.47 & 35.69 & 35.69 & 25.94 & 28.83 & 29.54 & 30.20 \\
        \toprule
        Exp 2.1 & Keyboard Augmentation             & 34.83 {\color{blue} \makebox[0pt][l]{$\uparrow$} } & 35.69 {\color{blue} \makebox[0pt][l]{$\uparrow$} } & 36.13 {\color{blue} \makebox[0pt][l]{$\uparrow$} } & 36.69 {\color{blue} \makebox[0pt][l]{$\uparrow$} } & 27.01 {\color{blue} \makebox[0pt][l]{$\uparrow$} } & 27.87 {\color{red} \makebox[0pt][l]{$\downarrow$} } & 28.21 {\color{red} \makebox[0pt][l]{$\downarrow$} } & 28.43 {\color{red} \makebox[0pt][l]{$\downarrow$} } \\
        Exp 2.2 & Swap Augmentation                 & 29.49 {\color{red} \makebox[0pt][l]{$\downarrow$} } & 30.54 {\color{red} \makebox[0pt][l]{$\downarrow$} } & 31.36 {\color{red} \makebox[0pt][l]{$\downarrow$} } & 32.07 {\color{red} \makebox[0pt][l]{$\downarrow$} } & 27.33 {\color{blue} \makebox[0pt][l]{$\uparrow$} } & 28.62 {\color{red} \makebox[0pt][l]{$\downarrow$} } & 29.13 {\color{red} \makebox[0pt][l]{$\downarrow$} } & 29.56 {\color{red} \makebox[0pt][l]{$\downarrow$} } \\
        Exp 2.3 & Delete Augmentation               & 28.59 {\color{red} \makebox[0pt][l]{$\downarrow$} } & 29.56 {\color{red} \makebox[0pt][l]{$\downarrow$} } & 30.01 {\color{red} \makebox[0pt][l]{$\downarrow$} } & 30.38 {\color{red} \makebox[0pt][l]{$\downarrow$} } & 27.81 {\color{blue} \makebox[0pt][l]{$\uparrow$} } & 28.95 {\color{blue} \makebox[0pt][l]{$\uparrow$} } & 29.16 {\color{red} \makebox[0pt][l]{$\downarrow$} } & 29.20 {\color{red} \makebox[0pt][l]{$\downarrow$} }\\
        Exp 2.4 & Spelling Augmentation             & 34.97 {\color{blue} \makebox[0pt][l]{$\uparrow$} } & 35.51 {\color{blue} \makebox[0pt][l]{$\uparrow$} } & 35.95 {\color{blue} \makebox[0pt][l]{$\uparrow$} } & 36.24 {\color{blue} \makebox[0pt][l]{$\uparrow$} } & 28.09 {\color{blue} \makebox[0pt][l]{$\uparrow$} } & 29.68 {\color{blue} \makebox[0pt][l]{$\uparrow$} } & 30.42 {\color{blue} \makebox[0pt][l]{$\uparrow$} } & 30.92 {\color{blue} \makebox[0pt][l]{$\uparrow$} } \\
        Exp 2.5 & Synonym Replacement               & 34.77 {\color{blue} \makebox[0pt][l]{$\uparrow$} } & 35.95 {\color{blue} \makebox[0pt][l]{$\uparrow$} } & 36.31 {\color{blue} \makebox[0pt][l]{$\uparrow$} } & 36.63 {\color{blue} \makebox[0pt][l]{$\uparrow$} } & 28.94 {\color{blue} \makebox[0pt][l]{$\uparrow$} } & 29.18 {\color{blue} \makebox[0pt][l]{$\uparrow$} } & 29.84 {\color{blue} \makebox[0pt][l]{$\uparrow$} } & 30.66 {\color{blue} \makebox[0pt][l]{$\uparrow$} } \\
        Exp 2.6 & Context Replacement               & 24.89 {\color{red} \makebox[0pt][l]{$\downarrow$} } & 26.04 {\color{red} \makebox[0pt][l]{$\downarrow$} } & 27.04 {\color{red} \makebox[0pt][l]{$\downarrow$} } & 27.60 {\color{red} \makebox[0pt][l]{$\downarrow$} } & 26.98 {\color{blue} \makebox[0pt][l]{$\uparrow$} } & 26.95 {\color{red} \makebox[0pt][l]{$\downarrow$} } & 28.03 {\color{red} \makebox[0pt][l]{$\downarrow$} } & 28.06 {\color{red} \makebox[0pt][l]{$\downarrow$} } \\
        Exp 2.7 & DAGA \citep{ding-etal-2020-daga}  & 32.75 {\color{red} \makebox[0pt][l]{$\downarrow$} } & 33.62 {\color{red} \makebox[0pt][l]{$\downarrow$} } & 34.17 {\color{red} \makebox[0pt][l]{$\downarrow$} } & 34.32 {\color{red} \makebox[0pt][l]{$\downarrow$} } & 28.57 {\color{blue} \makebox[0pt][l]{$\uparrow$} } & 29.29 {\color{blue} \makebox[0pt][l]{$\uparrow$} } & 29.95 {\color{blue} \makebox[0pt][l]{$\uparrow$} } & 30.54 {\color{blue} \makebox[0pt][l]{$\uparrow$} } \\
        \toprule
        Exp 2.8 & Ours (Domain Mapping)   & \bf 43.19 {\color{blue} \makebox[0pt][l]{$\uparrow$} } & \bf 43.85 {\color{blue} \makebox[0pt][l]{$\uparrow$} } & \bf 44.29 {\color{blue} \makebox[0pt][l]{$\uparrow$} } & \bf 44.82 {\color{blue} \makebox[0pt][l]{$\uparrow$} } & \bf 35.98 {\color{blue} \makebox[0pt][l]{$\uparrow$} } & \bf 38.33 {\color{blue} \makebox[0pt][l]{$\uparrow$} } & \bf 39.55 {\color{blue} \makebox[0pt][l]{$\uparrow$} } & \bf 40.84 {\color{blue} \makebox[0pt][l]{$\uparrow$} } \\
        \toprule
        \end{tabular}
    }
    \caption{Comparison of our proposed cross-domain mapping method with previous data augmentation method for NER task. Scores are calculated with the F1 metric. The best score for each column is in \textbf{bold}.}
    \label{ref: data_augmentation_results}
\end{table*}
Here we describe the experimental settings of the sequence labeling model for the NER experiment and report the evaluation results.

\paragraph{Sequence Labeling Model}
We fine-tune a BERT \citep{devlin-etal-2019-bert} model to evaluate our cross-domain mapping method on the NER task. BERT is pre-trained with masked language modeling and next sentence prediction objectives on the text from the general domain. We use BERT as the base model because it is capable of generating contextualized word representations and achieving high performances across many NLP tasks. We adapt a linear layer on top of BERT encoder to classify each token into pre-defined entity types. The hyper-parameters are described in Appendix \ref{sec: experimental_settings}.

\paragraph{Results}
To evaluate the quality of the generated data, we conduct experiments on the NER task with ten different domain pairs. The results are shown in Table \ref{ref: named_entity_recognition_results}. For each domain pair, we consider three experiments: 1) \textit{source domain}: train the model on the data from source domain as lower bound; 2) \textit{target domain}: train the model on the data from the target domain as upper bound; and 3) \textit{Gen}: train the model on the generated data combined with the data from the source domain. Based on the results, we observe that, when the patterns of text in the source and target domain are quite close (\textit{NW} as source domain), the improvement is quite limited or even no improvement. In most of the experiments with \textit{NW} as source domain, the improvement is less than 1\% F1 score. In the experiment of \textit{NW $\rightarrow$ BC}, we can see the performance becomes lower when we combine the generated data with the data from the source domain as training data. We suspect that this is because the discriminator in our model cannot distinguish which domain the latent vectors come from. For this reason, the model cannot generate meaningful intermediate representations and thus result in lower performances. However, when the patterns are dissimilar (\textit{SM} as source domain), \textit{Gen} can outperform the lower bound by up to 18.93\% F1 score, indicating that the model has a good understanding of the textual pattern from each domain, and that the textual pattern of the generated data are more similar to the data from the target domain than the data from the source domain.

\paragraph{Comparison with Previous Work}
To compare our proposed method with previous methods for data augmentation on NER, we augment the training data from the source domain (i.e, generate synthetic data and combine it with original training data) as the training set. The validation set and test set are from the target domain. We first establish a baseline Exp 2.0 \textit{Baseline (No Augmentation)} without using any data augmentation technique. Then we consider 7 different methods, including 1) Exp 2.1 \textit{Keyboard Augmentation}: randomly replace characters based on the keyboard distance, 2) Exp 2.2 \textit{Swap Augmentation}: randomly swap characters within each word, 3) Exp 2.3 \textit{Delete Augmentation}: delete characters randomly, 4) Exp 2.4 \textit{Spelling Augmentation}: substitute word by spelling mistake words dictionary, 5) Exp 2.5 \textit{Synonym Replacement}: substitute word by WordNet's \citep{Miller1995WordNetAL} synonym, 6) Exp 2.6 \textit{Context Replacement}: substitute words by BERT contextual word embeddings, and 7) Exp 2.7 \textit{DAGA} from \citet{ding-etal-2020-daga}. 

In Table \ref{ref: data_augmentation_results}, we compare our approach with the previous data augmentation method for the NER task by reporting the F1 score. We consider two different experiments: \textit{NW} $\rightarrow$ \textit{SM} and \textit{SM} $\rightarrow$ \textit{NW}. We augment the data from the source domain as training data. The validation data and test data are from the target domain. Based on the results, we observe that: 1) the improvement from traditional data augmentation (Exp 2.1 \textasciitilde Exp 2.6) is quite marginal. Only three of them can outperform the baseline in both \textit{NW} $\rightarrow$ \textit{SM} and \textit{SM} $\rightarrow$ \textit{NW} and the performance gain is from 0.34\% \textasciitilde 1\% F1 score; 2) data augmentation techniques are effective when we only have a small number of training samples. For example, in \textit{SM} $\rightarrow$ \textit{NW}, when using only 1K training samples, all methods can outperform the baseline. However, when using 4K training samples, only three of them can outperform the baseline;  3) Simply learning the textual patterns (Exp 2.7) in each domain may not always result in a good performance. When the size of training data is quite limited, the model struggles to learn the textual patterns and thus cannot achieve a good performance; and 4) transforming text from the source domain to the target domain is much powerful because, on average, it can outperform the baseline by 8.7\% and 10.1\% in two experiments, respectively.

\section{Analysis}
In this section, we take a step further towards exploring our approach in the context of the NER task in low-resource scenarios. Specifically, we investigate two crucial questions on data augmentation for both in-domain and out-of-domain datasets. 

\begin{figure}[t]
\centering
\subfigure{
    \begin{minipage}[t]{\linewidth}
    \centering
    \includegraphics[width=1\linewidth]{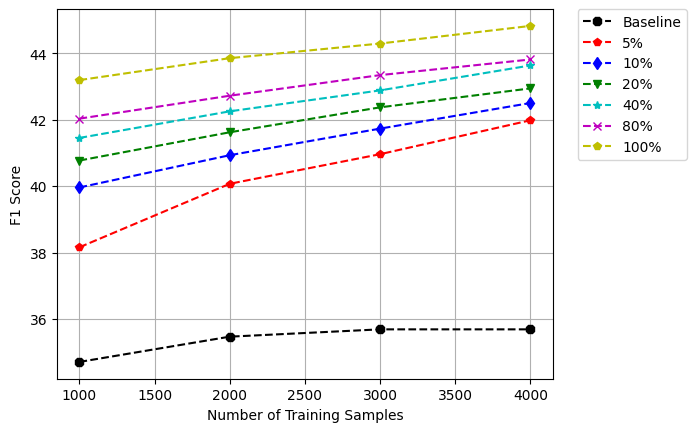}
    \end{minipage}
}
\caption{Model performance on the NER task with different amounts of target data and increasing amounts of augmented data for training. 
The test set is from \textit{SM}. Each curve shows model performance on NER when using different percentages of target data in the training set. The x-axis denotes the total number of training samples used. A 5\% means that out of all the training instances, only 5\% are coming from the target data, while the rest comes from the augmentation model. The y-axis denotes the F1 score for the NER task. }
\label{fig: low_resource_source}
\end{figure}

\paragraph{Q1: Does the model require a large number of training data from the target domain?}
To answer this question, we randomly select 5\%, 10\%, 20\%, 40\%, 80\%  of samples from the target domain as the train data to train our cross-domain autoencoder model. We consider \textit{NW} as the source domain and \textit{SM} as the target domain. After training the model, we generate samples from \textit{NW} to \textit{SM} and merge them into the training set from \textit{NW} as the new training set. Then we do evaluation on the test set from the \textit{SM}. We establish a baseline that only uses the data from \textit{NW} for training the model. Figure \ref{fig: low_resource_source} shows the results of model performance on the NER task. From this figure, we can see that our method can consistently achieve higher F1 scores than the baseline. Even if there is only 5\% training samples from the target domain, our model can still achieve significant improvements, averaging 4.81\% over the baseline.

\begin{figure}[t]
\centering
\subfigure{
    \begin{minipage}[t]{\linewidth}
    \centering
    \includegraphics[width=1\linewidth]{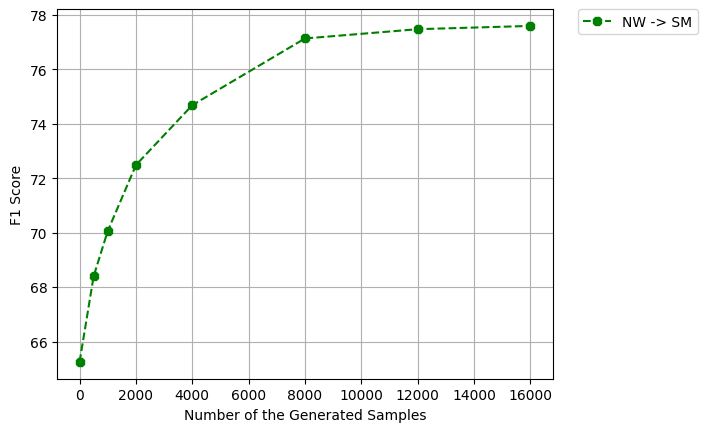}
    \end{minipage}
}
\caption{Model performance on the NER task with a fixed amount of target data and increasing amounts of augmented data for training. The training set starts with  500 instances from \textit{SM} (5\% of the training data in \textit{SM}) and the synthetic data generated from \textit{NW} to \textit{SM}. The test set is from \textit{SM}. The x-axis denotes the number of the generated samples that are combined with the data from \textit{SM} for training the NER model. The y-axis denotes the F1 score for the NER task. }
\label{fig: low_resource_target}
\end{figure}

\paragraph{Q2: Can we generate enough data to reach competitive results in the target domain?}
We train our cross-domain autoencoder model with all the data from \textit{NW} and only 5\% data (totally 500 samples) from \textit{SM}. Then we generate synthetic data from \textit{NW} to \textit{SM}. After that, we do evaluation on the NER task by combining 5\% data from \textit{SM} and different numbers of training samples as training data. Figure \ref{fig: low_resource_target} shows the F1 scores achieved by sequence labeling model. With 5\% data from \textit{SM}, the model can only reach 65.25\% F1 score. When we add more generated samples, the model can reach up to 77.59\% F1 score, pushing the model performance by 12.34\%.

\section{Discussion and Limitation}
In this work, we explore how to leverage existing data from high-resource domains to augment the data in low-resource domains. We introduce a cross-domain autoencoder model that captures the textual patterns in each domain and transforms the data between different domains. However, there are several limitations: 1) \textit{the maximum lengths of the generated sequences}: at each iteration, the maximum lengths of the generated sequences $\tilde{x}_{tgt}$ and $\tilde{x}_{src}$ are set to the same as original input sequences $x_{src}$ and $x_{tgt}$, respectively. This is not an ideal case because, intuitively, a short sentence in the source domain may correspond to a long sentence in the target domain. Fixing the maximum lengths of the generated sequences may hurt the model on capturing the semantics of original input sequences and result in a lower quality of reconstruction. 2) \textit{unparalleled datasets}: in our experiments, the datasets $D_{src}$ and $D_{tgt}$ are unparalleled, which means the sentences $x_{src}$ and $x_{tgt}$ do not correspond to each other. When we generate sequences from one domain to another, there is no guidance and thus we cannot control the quality of the generated sequences $\tilde{x}_{tgt}$ and $\tilde{x}_{src}$. 

Although our proposed method can not outperform the upper bound (i.e, training the model on the data from the target domain), it can achieve a significant improvement than only using the data from the source domain, providing a strong lower bound baseline for semi-supervised learning.

\section{Conclusion}
In this work, we present a novel neural architecture for data augmentation where the model learns to transform data from a high resource scenario to data that resembles that of a low resource domain. By training the model on reconstruction loss, it can extract the textual patterns in each domain and learn a feature space where both domains are aligned. We show the effectiveness of our proposed method by evaluating a model trained on the augmented data for NER, concluding that transforming text to low-resource domains is more powerful than only using the data from high-resource domains. Our future work includes three directions: i) explore how to embed more supervision about forcing a better alignment in the latent space, ii) design a strategy to control the quality of the generated sequences, and iii) generalize our method to other NLP tasks such as text classification. 

\section*{Acknowledgements}
This work was partially supported by the National Science Foundation (NSF) under grant \#1910192. We would like to thank the members from the RiTUAL lab at the University of Houston for their invaluable feedback. We also thank the anonymous EMNLP reviewers for their valuable suggestions.

\bibliography{anthology,custom}
\bibliographystyle{acl_natbib}

\appendix

\section{Experimental Settings}
\label{sec: experimental_settings}
This section describes the hyper-parameters for both cross-domain autoencoder model and sequence labeling model.

\paragraph{Cross-domain Autoencoder}
For cross-domain autoencoder, the embedding size is 512. The hidden state sizes of the LSTM and linear layer are set to 1024 and 300, respectively. The probability of word dropout is set as 0.1 to inject noise into input sequences. We use Adam \citep{Kingma2015AdamAM} as the optimizer with an initial learning rate of 5e-4 for both encoder and decoder. For the discriminator, we use RMSprop as the optimizer with initial learning rate 5e-4. The batch size is 32 and the number of training epochs is set to 50. We apply gradient clipping \citep{Pascanu2013OnTD} of 5 and the dropout rate is 0.5. In the first training phase, $\lambda_{1}$, $\lambda_{2}$, and $\lambda_{3}$ are set to 1, 0, and 10, respectively. In the second training phase, we change $\lambda_{2}$ to 1.

\paragraph{Sequence Labeling Model}
For sequence labeling model, the dropout rate is set to 0.1. We use AdamW \citep{Loshchilov2019DecoupledWD} as the optimizer with initial learning rate 5e-5 and weight decay is set to 0.01. The batch size is 32 and the number of training epochs is 20.

\section{Domain Similarity}
\label{sec: domain_similarity}
This section empirically analyzes the performance gains obtained by training models with synthetic data generated by our method. To this end, we analyze the data from the source and target domain, as well as the generated data. For this analysis, we consider two sets: train and test. The training set is used to directly update model parameters while the test set provides an unbiased evaluation of the final model. Entities that only appear in the training set are defined as non-overlapping entities, and those that appear in both the training set and the test set are defined as overlapping entities. The domain similarity is then defined as the percentage of overlap entities among all entities. As shown in Table \ref{ref: data_statistics}, the data from \textit{NW} is more similar to test data as it can achieve 10.90\% \textasciitilde 23.57\% domain similarity with the test data from other domains. Instead, the data from \textit{SM} can only achieve 0.22\% \textasciitilde 2.96\% domain similarity, indicating that the data from \textit{NW} is more close the data from other domains. Furthermore, the training data from the source domain achieves the lowest similarity while the training data from the target domain can always achieve the highest similarity. On average, our methods can improve the domain similarity by 8.25\%, which illustrates that the generated data can provide more diverse contexts for the entities that appear in the test set.
\begin{table}[t]
\small
    \centering
    \renewcommand{\arraystretch}{1.2}
    \resizebox{\linewidth}{!}{
    \begin{tabular}{l|l|rrr}
    \toprule
    \multirow{2}{*}{\bf Domain Pair} & \multirow{2}{*}{\bf Data} & \multicolumn{3}{c}{\bf Domain Similarity}\\ 
    \cmidrule(lr){3-5}
    & & \textbf{Non-overlap} & \textbf{Overlap} & \textbf{Similarity \%}\\
    \toprule
    \toprule
    \multirow{3}{*}{NW $\rightarrow$ BC}& NW        & 63,666    & 14,056    & 18.08\\
                                        & Gen       & 11,901    &  7,216    & 37.75\\
                                        & BC        &  5,927    &  3,877    & 39.55\\
    \toprule
    \multirow{3}{*}{NW $\rightarrow$ BN}& NW        & 59,405    & 18,317    & 23.57\\
                                        & Gen       & 34,994    & 12,734    & 26.68\\
                                        & BN        & 27,764    &  5,437    & 55.43\\
    \toprule
    \multirow{3}{*}{NW $\rightarrow$ MZ}& NW        & 69,248    &  8,474    & 10.90\\
                                        & Gen       & 27,764    &  5,437    & 16.38\\
                                        & MZ        &  6,738    &  4,183    & 38.30\\
    \toprule
    \multirow{3}{*}{NW $\rightarrow$ WB}& NW        & 66,779    & 10,943    & 14.08\\
                                        & Gen       & 24,582    &  3,638    & 12.89\\
                                        & WB        &  5,578    &  2,718    & 32.76\\
    \toprule
    \multirow{3}{*}{NW $\rightarrow$ SM}& NW        & 76,347    &  1,375    & 1.77\\
                                        & Gen       & 14,347    &  2,895    & 16.79\\
                                        & SM        &  4,970    &  1,418    & 22.20\\
    \toprule
    \toprule
    \multirow{3}{*}{SM $\rightarrow$ BC}& SM        &  6,250    &    138    & 2.16\\
                                        & Gen       & 10,996    &    978    & 8.17\\
                                        & BC        &  5,927    &  3,877    & 39.55\\
    \toprule
    \multirow{3}{*}{SM $\rightarrow$ BN}& SM        &  6,199    &    189    & 2.96\\
                                        & Gen       & 10,302    &  2,172    & 17.41\\
                                        & BN        &  8,807    & 10,954    & 55.43\\
    \toprule
    \multirow{3}{*}{SM $\rightarrow$ MZ}& SM        &  6,374    &     14    & 0.22\\
                                        & Gen       & 19,392    &    769    & 3.81\\
                                        & MZ        &  6,738    &  4,183    & 38.30\\
    \toprule
    \multirow{3}{*}{SM $\rightarrow$ NW}& SM        &  6,207    &    181    & 2.83\\
                                        & Gen       & 10,564    &  1,810    & 14.63\\
                                        & NW        & 46,954    & 30,768    & 39.59\\
    \toprule
    \multirow{3}{*}{SM $\rightarrow$ WB}& SM        &  6,342    &     46    & 0.72\\
                                        & Gen       &  8,951    &    498    & 5.27\\
                                        & WB        &  5,578    &  2,718    & 32.76\\
    \toprule
    \toprule
    \end{tabular}
    }
    \caption{The statistics of domain similarity. \textbf{Overlap} describes the set operations: Train $\cap$ Test. Notably, the overlap percentages of the data from the target domain are substantially higher than the ones from the source domain. \textbf{Similarity} refers to the percentage of total overlap entities out of the all entities (including both overlap and non-overlap entities).
    }
    \label{ref: data_statistics}
\end{table}

\end{document}